\documentclass{nologos}

\usepackage[utf8]{inputenc}
\usepackage{amsmath}
\usepackage{subfig}
\usepackage{graphicx}
\usepackage[rightcaption]{sidecap}
\usepackage{multirow}
\usepackage{makecell}
\usepackage{url}

{}
{}
{}

\begin{document}

\supertitle{Multiple Hypothesis Tracking Algorithm for Multi-Target Multi-Camera Tracking with Disjoint Views}

\title{Multiple Hypothesis Tracking Algorithm for Multi-Target Multi-Camera Tracking with Disjoint Views}

\author{\au{Kwangjin Yoon}, \au{Young-min Song}, \au{Moongu Jeon$^{\corr}$}}

\address{\add{}{Electrical Engineering and Computer Science, Gwangju Institute of Science and Technology, Cheomdangwagi-ro 123, Buk-gu, Gwangju, Republic of Korea}
\email{mgjeon@gist.ac.kr}}

\begin{abstract}
In this study, a multiple hypothesis tracking (MHT) algorithm for multi-target multi-camera tracking (MCT) with disjoint views is proposed. The authors' method forms track-hypothesis trees, and each branch of them represents a multi-camera track of a target that may move within a camera as well as move across cameras. Furthermore, multi-target tracking within a camera is performed simultaneously with the tree formation by manipulating a status of each track hypothesis. Each status represents three different stages of a multi-camera track: \textit{tracking}, \textit{searching}, and \textit{end-of-track}. The tracking status means targets are tracked by a single camera tracker. In the searching status, the disappeared targets are examined if they reappear in other cameras. The end-of-track status does the target exited the camera network due to its lengthy invisibility. These three status assists MHT to form the track-hypothesis trees for multi-camera tracking. Furthermore, they present a \textit{gating} technique for eliminating of unlikely observation-to-track association. In the experiments, they evaluate the proposed method using two datasets, DukeMTMC and NLPR\_MCT, which demonstrates that the proposed method outperforms the state-of-the-art method in terms of improvement of the accuracy. In addition, they show that the proposed method can operate in real-time and online.
\end{abstract}

\maketitle

\section{Introduction}

A large number of cameras recently have been deployed to cover wide area. Besides, tracking multiple targets in a camera network becomes an important and challenging problem in visual surveillance systems since in-person monitoring wide area is costly and needs a lot of effort. Hence, it is desirable to develop multi-target multi-camera tracking (MTMCT) algorithm.
 
In this paper, our goal is to develop an algorithm that can track multiple targets (especially for pedestrians in this work) in a camera network. The targets may move within a camera or move to another camera and the coverage of each camera does not overlap. To achieve this goal, we need to solve both single camera tracking (SCT) and multi-camera tracking (MCT). There has been great amount of effort made to SCT whereas relatively smaller amount of effort has been done for MCT with disjoint views. Moreover, most MCT approaches\cite{ chen2014object, javed2003tracking, wang2014distributed} only focus on tracking targets across cameras by assuming solved SCT in advance; thus, jointly tracking multiple targets in both within and across cameras still remains to be explored much further\cite{tesfaye2017multi}.

The proposed MHT algorithm tracks targets across cameras by maintaining the identities of observations which are obtained by solving SCT that tracks targets in within-camera. Thus, our method jointly tracks targets in both within and across cameras. In this work, we adopt the real-time and online method\cite{ristani2014tracking} to produce observations by tracking multiple targets in within-camera. These observations obtained from each camera are fed into the proposed MHT algorithm which solves MCT problem. The proposed MHT algorithm forms track-hypothesis trees with obtained observations either by adding a child node to hypothesis tree, which describes the association between an observation and an existing track hypothesis, or by creating a new tree with one root node indicating an observation, which describes the initiation of a new multi-camera track. Each branch in track-hypothesis trees represents different across camera data association result (i.e., a multi-camera track). To work in concert with SCT, every node in track-hypothesis trees designates certain observation and all leaf nodes have a status. There are three statuses for the proposed MHT and each of which represents a different stage of a multi-camera track, \textit{tracking}, \textit{searching}, and \textit{end-of-track}. With the status, the MHT can form the track-hypothesis trees while simultaneously solving SCT to produce observations. Then it selects the best set of track hypotheses as the multi-camera tracks from the track-hypothesis trees. Furthermore, we propose \textit{gating} mechanism to eliminate unlikely observation-to-track pairing; this also prevents track-hypothesis trees from unnecessary growth. We propose two gating mechanisms, speed gating and temporal gating in order to deal different tracking scenarios (tracking targets on the ground plane or image plane).

For the appearance feature of an observation, we used simple averaged color histogram as an appearance model after Convolutional Pose Machine\cite{wei2016cpm} is applied to an image patch of a person in order to capture the pose variation. The experimental results shows that our method achieves state-of-the-art performance on DukeMTMC dataset and performs comparable to the state-of-the-art method on NLPR\_MCT dataset. Furthermore, we demonstrate that proposed method is able to operate in real-time with real-time SCT in Section \ref{sec:realtime}.

The remainder of this paper organized as follows. In Section \ref{sec:relatedworks}, we review relevant previous works. The detailed explanation of proposed method is given in Section \ref{sec:method}. Section \ref{sec:mht-tree} describes how the proposed MHT forms track-hypothesis trees while it simultaneously works with SCT. The proposed gating mechanism is explained in Section \ref{sec:gating}. In Section \ref{sec:exp}, we report experiment results with conducted on DukeMTMC and NLPR\_MCT datasets. Finally, we conclude the paper in Section \ref{sec:conclude}.

\section{Related Works}\label{sec:relatedworks}

Single camera tracking (SCT), which tracks multiple targets in a single scene, is also called multi-object tracking (MOT). Many approaches have been proposed to improve the MOT. Track-by-detection, which optimizes a global objective function over many frames have emerged as a powerful MOT algorithm in recent years\cite{wang2016tracking}. Network flow-based methods are successful approaches in track-by-detection techniques\cite{berclaz2011multiple, zhang2008global, pirsiavash2011globally}. These methods efficiently optimize their objective function using the push-relabel method\cite{zhang2008global} and successive shortest path algorithms\cite{berclaz2011multiple, pirsiavash2011globally}. However, the pairwise terms in network flow formulation are restrictive in representing higher-order motion models, e.g., linear motion model and constant velocity model\cite{collins2012multitarget}. In contrast, formalizing multi-object tracking with multidimensional assignment (MDA) problem produces more general representations of computed trajectories since MDA can exploit the higher-order information\cite{collins2012multitarget,mht-rv}. Solutions for MDA are MHT\cite{mht-old,mht-rv,papageorgiou2009maximum} and Markov Chain Monte Carlo (MCMC) data association\cite{oh2009markov}. While MCMC data association exploits the stochastic method, MHT searches the solution space deterministic way.

Multiple Hypothesis Tracking(MHT) was first presented in \cite{reid1979algorithm} and is regarded as one of the earliest successful algorithm for visual tracking. MHT maintains all track hypotheses by building track-hypothesis trees whose branch represent a possible data association result(a track hypothesis). The probability of a track hypothesis is computed by evaluating the quality of data association result the branch had. An ambiguity of data association which occurs due to either short occlusion or missed detection does not usually matter for MHT since the best hypothesis is computed with higher-order data association information and entire track hypotheses. In this paper, we applied MHT to solve the multi-camera tracking problem. 

Multi-camera tracking aims to establish target correspondences among observations obtained from multiple cameras so as to achieve consistent target labelling across all cameras in the camera network \cite{tesfaye2017multi}. Earlier research works in MCT only try to address tracking targets across cameras, assuming solved SCT. However, researchers have argued recently that assumptions of availability of intra-camera tracks are unrealistic \cite{wang2014distributed}. Therefore, solving MCT problem by simultaneously treating problem of SCT seems to address more realistic problem. Y.T Tesfaye \etal \cite{tesfaye2017multi} proposed a constrained dominant set clustering (CDSC) based framework that utilizes a three layers hierarchical approach, where SCT problem is solved using first two layers, and later in the third layer MCT problem is solved by merging tracks of the same person across different cameras. In this paper, we also solve the problem of across camera data association(MCT) by the proposed MHT while SCT is simultaneously treated by real-time multi-object tracker such as \cite{ristani2014tracking,song2016online}.

Multi-camera tracking with disjoint views is a challenging problem, since illumination and pose of a camera changes across cameras as well as a track discontinues owing to the blind area of camera network or miss detections. Some MCT methods try to relax the variation in illumination using appearance cue. O. Javed \etal \cite{javed2008modeling} suggested the brightness transfer function to deal with illumination change as a target moves across cameras. B.J. Prosser \etal \cite{prosser2008multi} used Cumulative Brightness Transfer Function that can learn from very sparse training set and  A. Gilbert \etal \cite{gilbert2006tracking} proposed incremental learning method to model the color variations. S. Srivastaba \etal \cite{srivastava2011color} suggested color correction method for MCT in order to achieve color consistency for each target across cameras. Recent methods have used not only the appearance cue but also the space-time cue to improve the performance of MCT. C. Kuo \etal \cite{kuo2010inter} first learned an appearance model for each target and combined it with space-time information. They show that their proposed combined model improved the performance of across-camera tracking. S. Zhang \etal \cite{zhang2015tracking} tracked multiple interacting targets by formulating the problem into network flow problem. They identified the group merge and split events using space-time relationship among targets. In \cite{chen2014object}, they learn across-camera transfer model using both space-time and appearance cues. They designed space-time transfer model as normal distribution and learned the parameters using cross-correlation function. For appearance transfer model, they used color transfer method to capture color variations across the cameras. Ergys \etal solved SCT\cite{ristani2014tracking} by transforming the problem into graph partitioning problem and extending their approach to multi-camera tracking with disjoint views. There are some approaches to solve MCT problem using person re-identification \cite{rev-pp1, rev-pp2}. L. Chen \etal \cite{rev-pp1} proposed a deep neural network architecture composed of convolutional neural network (CNN) and recurrent neural network (RNN) that can jointly exploit the spatial and temporal information for the video-based person re-identification. C.W. Wu \etal \cite{rev-pp2} designed a track-based multi-camera tracking (T-MCT) framework with person re-identification algorithms. Their method found multi-camera tracks using re-identification algorithms as both the feature extractor of an object and the distance metric between two object. They also proposed new evaluation metrics for MCT to report the performance of T-MCT with various re-identification algorithms.

\section{Method}\label{sec:method}

For a multi-camera multi-target tracking system with disjoint views, the set $S=\{s_{c}\}_{c=1:C}$ is the set of all cameras in the camera network, where $C$ is the number of cameras. Let $K$ denote the most recent time. Single camera tracker of each camera generates observations $o_i$ so that they form a set of observations $O = \{o_i\}_{i=1:N}$ where $N$ is the total number of observations observed until the most recent time $K$. An observation contains all the information about the target while it was being tracked by a single camera tracker. Specifically, the $i$-th observation $o_i =\{A_i,X_i,\pi_i\}$ consists of the appearance feature $A_i$, the track $X_i$ and the camera in which it appears $\pi_i \in S$. The track $X_i$ is a collection of all track histories of observation $o_i$ observed in $\pi_i$, i.e.\ $X_i = \{\mathbf{x}_i^l\}_{l=1:|X_i|}$ where $|X_i|$ represents the length of track of $o_i$ recorded until time $K$ and $\mathbf{x}_i^l=(t_i^l, u_i^l, v_i^l, w_i^l, h_i^l, \mathbf{y}_i^l)$ refers the $l$-th track history of $o_i$ containing time stamp $t_i^l$, position $(u_i^l,v_i^l)$ and size $(w_i^l,h_i^l)$ in image plane of camera $\pi_i$, and $\mathbf{y}_i^l = (x_i^l, y_i^l)$ is a position on ground plane. Note that  depending on the scenario, $\mathbf{x}_i^l$ might not contain $\mathbf{y}_i^l$. With this observation set, the MCT system outputs a set of multi-camera tracks, $T=\{T_j\}_{j=1:|T|}$, where $|T|$ is the size of set $T$ and $T_j=\{o_{I_{j}}\}_{I_j\subseteq \{1:N\}}$ refers to the $j$-th multi-camera track which consists of observations, i.e.\ a number of observations(or single observation) which have the same identity composes a multi-camera track. The $I_{j}$ is an index set whose elements indicate elements of set $O$; hence, $o_{I_{j}}$ is a subset of $O$. Here, we introduce a notation $\{I_j^l\}_{l=1:|I_j|}$ enumerating the set $I_j$ and $I_j^l$ (the $l$-th element of $I_j$) indicates an observation that is also the $l$-th observation of $T_j$. Thus, we can write $T_j^l = o_{I_j^l} = o_i$, which means that the $l$-th observation of the multi-camera track $T_j$ is the $o_i$ in the set $O$. Therefore, $|I_j|$ is not only the number of elements in $I_j$ but also the number of observations that the $T_j$ had. Finally, for all observations and all multi-camera tracks there is the constraint that one observation must belong to a unique multi-camera track such that:

\begin{equation}\label{eq:set-constraint}
    T_j \cap T_k = \emptyset, \forall T_j,T_k \in T, j \ne k .
\end{equation}
i.e.\ all tracks in the set $T$ do not conflict each other.

\subsection{Multiple Hypothesis Tracking for MCT}\label{sec:mht-tree}

\begin{figure*}[t]
\centering
\subfloat[Observations in camera network]{\label{fig:trackform:a}\includegraphics[height=1.25in]{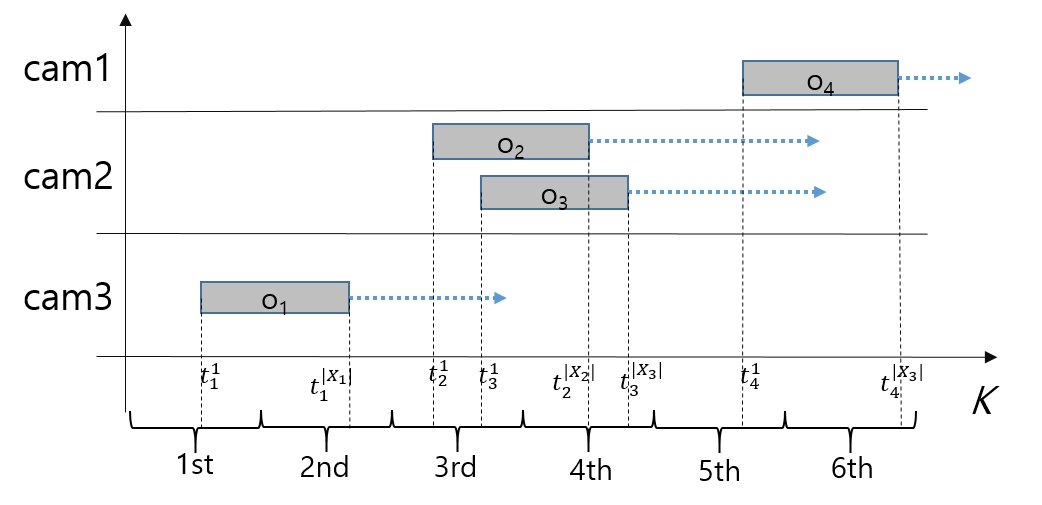}} \qquad
\subfloat[First scan]{\label{fig:trackform:b}\includegraphics[height=0.95in]{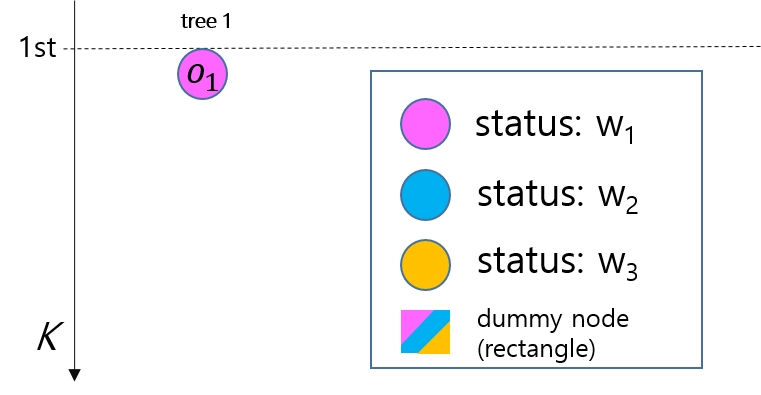}} \qquad
\subfloat[Third scan]{\label{fig:trackform:c}\includegraphics[height=0.95in]{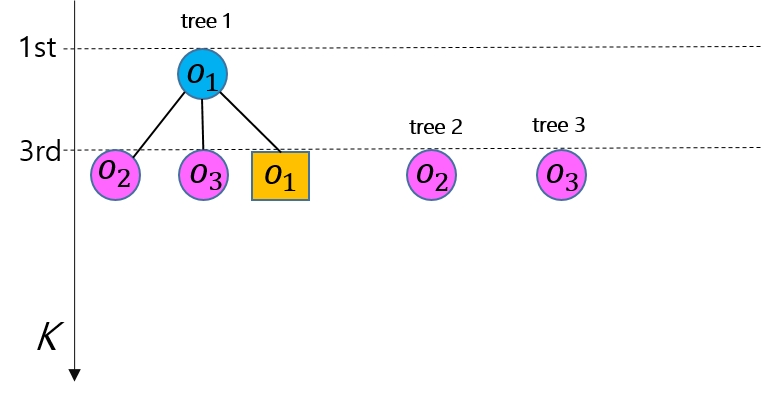}}

\subfloat[Fourth scan]{\label{fig:trackform:d}\includegraphics[height=0.95in]{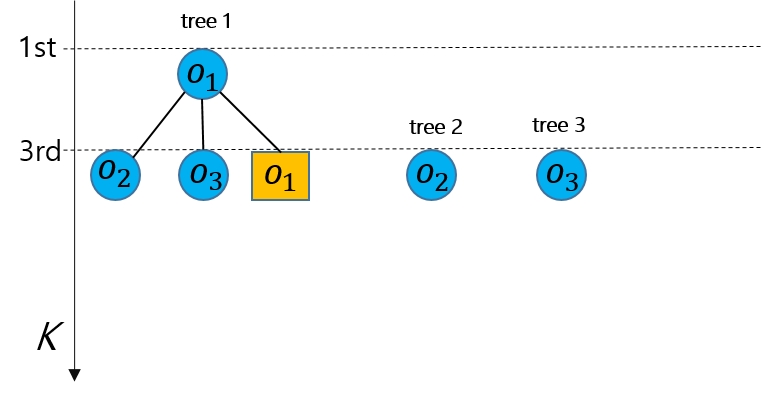}} \qquad
\subfloat[Sixth scan]{\label{fig:trackform:e}\includegraphics[height=0.95in]{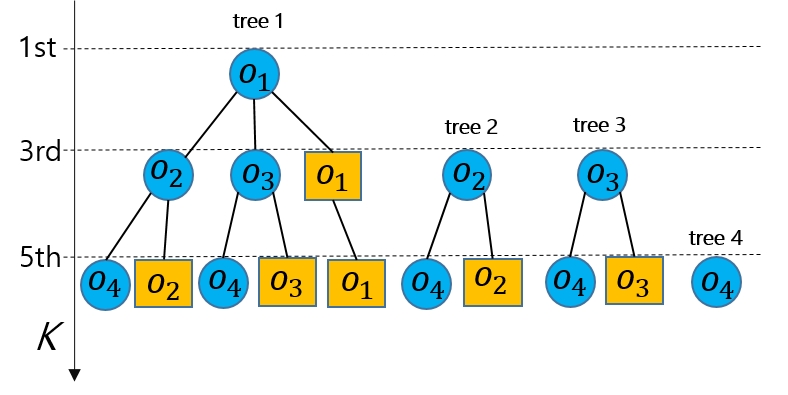}} \qquad
\subfloat[$N$-scan pruning]{\label{fig:trackform:f}\includegraphics[height=0.95in]{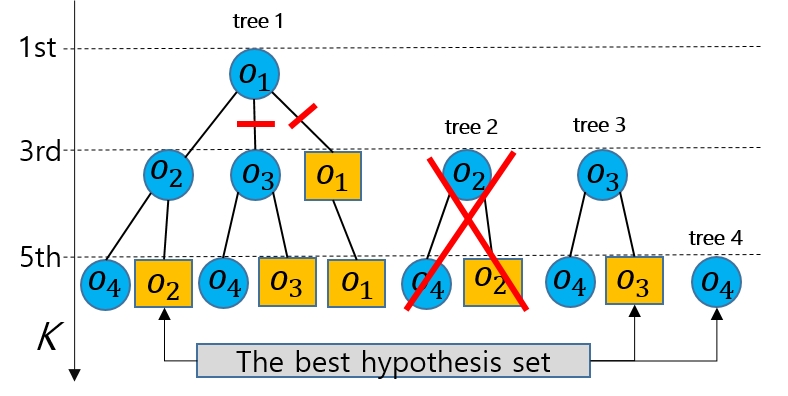}}
\caption{ Example of track-hypothesis tree formation: \protect\subref{fig:trackform:a} Four gray rectangles represent observations and their widths show the length of track. The ends of the blue arrows at the back of each observation represent their maximum gating time. The vertical position of an observation refers to the camera in which it appeared (in this example there were three cameras). The horizontal position of an observation shows the time information of the track. The under-braces at the bottom show the time intervals of each scan. \protect\subref{fig:trackform:b},\protect\subref{fig:trackform:c},\protect\subref{fig:trackform:d},\protect\subref{fig:trackform:e} show the results of track-hypothesis tree formation after the first, third, fourth, and sixth scan, respectively. The ordinals on the vertical axis represent the scan time where corresponding nodes were formed. In addition, a node in the tree refers to an observation. \protect\subref{fig:trackform:f} Result of $N$-scan pruning is shown after the best hypothesis set is computed ($N=2$). The dummy nodes are marked with rectangles and can be any of the three status. For a more detailed explanation, refer to Section~\protect\ref{sec:mht-tree}. }
\label{fig:trackform}
\end{figure*}

In this section, we introduce how a track-hypothesis tree is formed for a multi-camera tracking system. The tree of our method maintains multi-camera tracks by initiating, terminating, updating with new observations. A node of the tree represents an observation that is generated by SCT. All branches of the tree represent all possible hypotheses that originate from a single observation or root node. A key strategy of MHT is to delay data association decisions by keeping multiple hypotheses active until data association ambiguities are resolved\cite{mht-rv}. As new observations are received, MHT\cite{blackman-book} forms new trees to initiate tracks for each new observation. Then existing tracks are updated with new observations that were within the gate. Moreover, all existing tracks are updated with dummy observations in order to describe the hypothesis that they are not updated with any current observation(missing detection). Consequently, the number of track hypotheses continues to expand and many of the tracks are inconsistent since the same observations are used for more than one track.

\begin{figure*}[t]
\centering
\subfloat[Long scan time]{\label{fig:scantime:a}\includegraphics[width=0.34\textwidth]{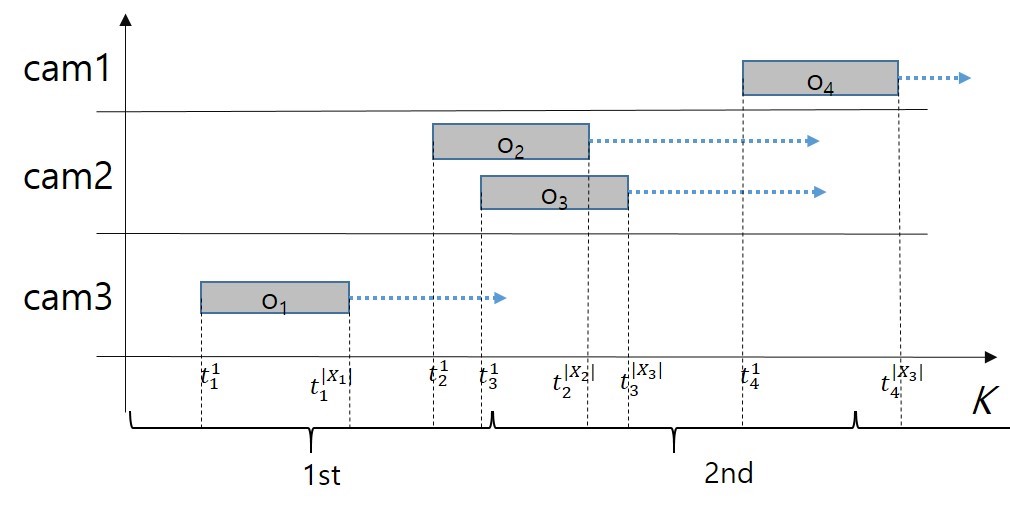}} \qquad
\subfloat[Short scan time]{\label{fig:scantime:b}\includegraphics[width=0.34\textwidth]{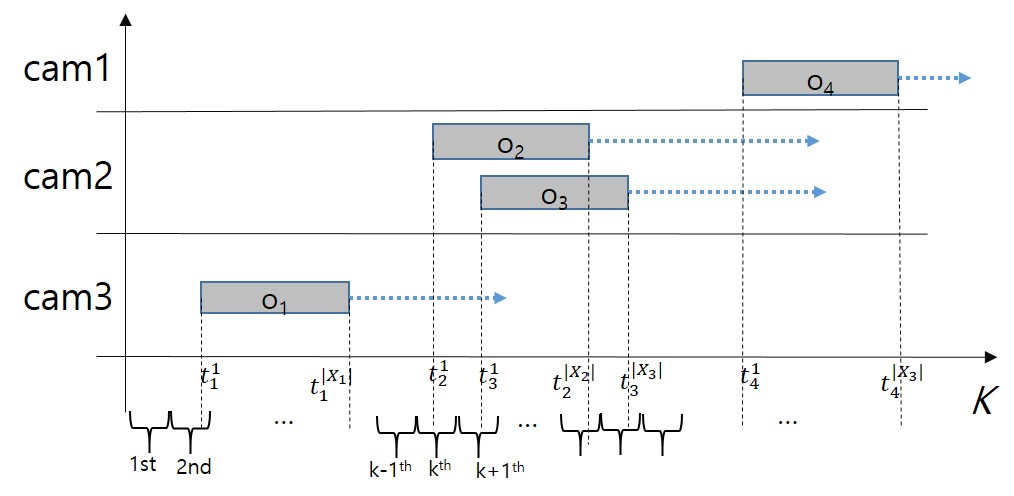}}
\caption{ \protect\subref{fig:scantime:a} An example of long scan time. The system cannot generate hypotheses associating observation $o_1$ with $o_2$ as well as $o_1$ with $o_3$. \protect\subref{fig:scantime:b} An example of short scan time. The existing track hypotheses will be updated with new and dummy observations at $k$-th and $(k+1)$-th scan. Note that the dummy node is not appended at ($k-1$)-th scan where no new observation was received. Figure \protect\ref{fig:trackform:a} explains the above example.}
\label{fig:scantime}
\end{figure*}

In the tracking literature, a \textit{scan} is the time interval and sensor FoV (field of view) where observations are collected\cite{blackman-book}. With previous MHT algorithms for vision based target tracking systems\cite{mht-rv, mht-old}, images were scanned frame-by-frame to gather observations using a feature detector such as person detector\cite{felzenszwalb2010object} and corner detector\cite{lucas1981iterative}. Hence, the depth of their track-hypothesis trees grow with every frame. However, unlike their approach, we gather observations by scanning the entire camera network within a fixed amount of time; consequently, our tree is extended after a scan. Setting an appropriate time interval for one scan is important, because a scan should not contain multi-camera tracks. For example, if the interval is long enough to have observations that could form a multi-camera tracks; then the system loses the chance to associate them correctly(Figure~\ref{fig:scantime:a}). On the other hand, too-short time interval for a scan leads to increased computational overhead as well as deepened track-hypothesis tree due to the frequent update of trees. The amount of time for a scan generally depends on the datasets. Furthermore, to prevent the trees from growing meaninglessly by appending only dummy node to all branches, trees are extended with dummy only when a scan contains new observations (Figure~\ref{fig:scantime:b}). This enhances the efficiency of tree formation if pedestrians enter into the camera network sparsely.

Now we introduce the three statuses $\{w_1,w_2,w_3\}$. Every leaf node of our track-hypothesis trees should have a status in order that a branch, or a track hypothesis, has a status. Note that intermediate nodes have no effect on the status. The first status, $w_1$, is \textit{tracking} status, meaning that the target is being tracked by a single camera tracker. Therefore, all tracks are initiated with status $w_1$ and the track hypothesis in this status is not updated with new observations. If the target disappears from a camera, the status of leaves which refer to the target changes to \textit{searching}, $w_2$. A leaf node with status $w_2$ is updated with new observations that  satisfy the gating condition. The last status, $w_3$, is \textit{end-of-track}, which means that the target exits the camera network due to its invisibility for a long period of time. A branch with this status will not be updated with new observations except with dummy observations. A leaf with status $w_2$ changes to status $w_3$ by checking the elapsed time from when the leaf began status $w_2$ to most recent time $K$(refer to Section~\ref{sec:gating}). Note that only leaf nodes in status $w_2$ can be updated with new observations by appending a node referring to a new observation as its children. Otherwise, a leaf node in either status $w_1$ or $w_3$ only appends a dummy observation indicating the same observation as its parent after the scan that receives any new observations. Thus, once the status of a track hypothesis has changed to $w_3$, it can not revert to either $w_1$ or $w_2$. From here, we explain how we form the tree using the example in Figure~\ref{fig:trackform:a}. In Figure~\ref{fig:trackform:b}, after the first scan, a tree(\textit{tree 1}) with a node indicating observation $o_1$ is formed to make a track hypothesis that $o_1$ initiates a new multi-camera track. After the second scan, the status of the leaf node of \textit{tree 1} changes to $w_2$ because $o_1$ is no longer seen by Camera 1. In the third scan, two new observations, $o_2$ and $o_3$, are received; hence, two trees, \textit{tree 2} and \textit{tree 3}, are newly formed for $o_2$ and $o_3$, respectively. Then, the existing track hypothesis in status $w_2$ (the root node of \textit{tree 1}) is associated with new observations by appending nodes referring to $o_2$ and $o_3$, respectively. A dummy node is also added in order to describe the hypothesis that both $o_2$ and $o_3$ were not $o_1$(Figure~\ref{fig:trackform:c}). Note that we only consider the temporal gating scheme in this example for simplicity and the gating time of $o_1$ covers initiation time of both observations, $t_2^1$ and $t_3^1$. After the third scan, because the gating time of $o_1$ is expired, the status of leaf node referring to $o_1$ is changed to $w_3$ (rightmost leaf of \textit{tree 1} in Figure~\ref{fig:trackform:c}). Next, two targets exit from the Camera 2 in the fourth scan. The status of the leaf nodes related to the exited targets is changed. For $o_2$, the status of two leaves are changed to $w_2$ because $o_2$ occurred twice in the track-hypothesis trees(leftmost leaf node of \textit{tree 1} and root node of \textit{tree 2} in Figure~\ref{fig:trackform:d}). This works again for $o_3$. Because no new observation is received in this scan, trees are not extended with dummy nodes. Finally, Figure~\ref{fig:trackform:e} shows the result of track-hypothesis tree formation after the sixth scan. With these track trees, all possible data association hypotheses can be made. The first branch of \textit{tree 1} shows a multi-camera track with a hypothesis that describes that the observations $o_1, o_2,$ and $o_4$ have the same identity, while the last branch of \textit{tree 1} shows that the multi-camera track consists of only one observation, $o_1$(Figure~\ref{fig:trackform:e}). 

To compute the set that satisfy the constraint (\ref{eq:set-constraint}), each track hypothesis should manage the incompatible track lists. For example, in Figure~\ref{fig:trackform:e}, the incompatible tracks of the first branch of \textit{tree 1} are the track hypotheses that have observation $o_1, o_2$, and $o_4$, i.e.\ all the other branches in same the tree (because they at least share $o_1$), all track hypotheses in \textit{tree 2}(because of $o_2$ and $o_4$), the first branch of \textit{tree 3}(because of $o_4$), and the root node of \textit{tree 4}(because of $o_4$). Then the set of best track hypotheses is computed by solving a maximum weighted independent set problem\cite{mht-rv} which will be described in the Section \ref{sec:besthypo}.

\subsection{Gating}\label{sec:gating}

\textit{Gating} is a technique for eliminating unlikely observation-to-track pairings. In \cite{mht-rv, mht-old}, the spatial distance between the predicted location of an existing track and a newly received observation was used to determine whether to update an existing track with a new observation. They used the velocity of existing tracks to compute the predicted locations. If the distance between the predicted location and observation exceeds a pre-determined threshold, the track is not updated with that observation. However, for multi-camera tracking with disjoint views, predicting the location of re-appearance is very difficult because targets move through blinded areas for a long time. To resolve this problem, distance between observations is used instead of predicting the location. Note that, in this case, the world coordinates of any given track are known. Let $o_i$ be the last observation of an existing track and $o_j$ be the newly received observation. Then the following inequality checks the speed gating,
\begin{equation}\label{eq:speedgating}
\resizebox{0.9\linewidth}{!}{ $
    G_{speed}^{min} < \frac{\beta\|\mathbf{y}_j^1-\mathbf{y}_i^{|X_i|}\|_2 + (1-\beta)\|\mathbf{y}_j^1-\mathbf{y}_i^{|X_i|}\|_1}{t_j^1 - t_i^{|X_i|}} < G_{speed}^{max}$ }
\end{equation}
where $\mathbf{y}_i^l = (x_i^l, y_i^l)$ is the world coordinate of the $l$-th track of $o_i$ on ground plane, $G_{speed}^{min}$ is the minimum speed of a target, and $G_{speed}^{max}$ is the maximum speed of a target. These can either be set by the system designer or learned from training samples. This assumed that a target can not move faster (slower) than the threshold. $\beta$ is the control parameter of the distance metric between Euclidean and Manhattan distance, $0\leq\beta\leq1$,  which is also set by the system designer. This parameter is beneficial since we do not know what will transpire in blind areas. If an observation-to-track pair does not satisfy the inequality, then that pair will not be associated. A leaf node in status $w_2$ that refers to observation $o_i$ changes status to $w_3$ if specified time, $G_{i}^{end}$, has elapsed after when the leaf node began the status $w_2$. That is, if $K-t_i^{|X_i|} > G_{i}^{end}$, then the leaf node change status to $w_3$, where $K$ is the most recent time. $G_{i}^{end}$ is computed by $G_{i}^{end}=\sqrt{G_{area}}/\dot{y}_i$ where $G_{area}$ is the area of the ground plane of the camera network and $\dot{y}_i$ is the estimated speed of $o_i$.

In some situations, it is impossible to locate targets that are in tracking status($w_1$) due to the absence of calibration information and map information. In this case, we use temporal gating instead of speed gating. First, we estimate entry/exit points for each camera either by learning from training samples or by getting information from the system designer. Let $E=\{e_j\}_{j=1:|E|}$ be the set of entry/exit points, and $e_j$ is the $j$-th entry/exit point, where $|E|$ represents the total number of entry/exit points in the camera network. Then $\pi_i$ of $o_i$ has two elements, $\pi_i^{en}$ and $\pi_i^{ex}$, which represent the entry and exit point of the observation, respectively, i.e.\ $\pi_i = \{ \pi_i^{en}, \pi_i^{ex} \}$, and $\pi_i^{en}, \pi_i^{ex} \in E$. After that, we learned the transition matrix between each pairs of entry/exit points as well as the mean and standard deviation of transition time using training samples. Let $\Phi$ be the transition matrix, a square matrix with size $|E|\times|E|$. An element $\Phi_{i,j}$, $i$-th row and $j$-th column, is set to one if a transition from $e_i$ to $e_j$ exists. Otherwise, it is set to zero. For all $\Phi_{i,j}=1$, we learned the mean and standard deviation matrix of transition time with training samples. Then the temporal gating for the existing track whose leaf node designates $o_i$ and newly received observation $o_j$ checks the followings:
\begin{equation}\label{eq:timegating}
    G_{time}^{min}< t_j^1 - t_i^{|X_i|} < G_{time}^{max},\quad \Phi_{\pi_i^{ex},\pi_j^{en}}=1,
\end{equation}
where $G_{time}^{min}$ and $G_{time}^{max}$ are the minimum and maximum threshold for temporal gating, and which can be learned from the training samples. Note that middle-term of the inequality is always positive for MCT with disjoint view, otherwise its absolute value is needed. If the mean transition time between $\pi_i^{ex}$ and $\pi_j^{en}$ is $\mu$ and its standard deviation is $\sigma$, $G_{time}^{min}$ and $G_{time}^{max}$ could be set to $\mu-\alpha_1\sigma$ and $\mu+\alpha_2\sigma$, respectively, where $\alpha_1$, $\alpha_2$ are set by the system designer. If an observation-to-track pair does not satisfy the check, then that pair will not be associated. A leaf node in status $w_2$ referring to observation $o_i$ changes the status to $w_3$ if the time gap between last observed time and recent time $K$ is beyond the predetermined gating time, i.e.\ it will change the state to $w_3$ if $K-t_i^{|X_i|}$ is larger than $G^{end}$. In this case, $G^{end}$ is used for all observations. 

\subsection{Pruning}\label{pruning}

Since there is potential for a combinatory explosion in the number of track hypotheses that our MHT system could generate, \textit{pruning} the track-hypothesis tree is an essential task for MHT. We adopted the standard $N$-scan pruning technique\cite{blackman-book, mht-rv}. The standard $N$-scan pruning algorithm assumes that any ambiguity at $K$ is resolved by time $K+N$, i.e.\ it defines the number of frames to look ahead in order to resolve an ambiguity\cite{mht-old}. Note that in our case, $N$ refers not to the number of frames but to the number of scans since our trees grow after the scan where any new observation is received. An example of $N$-scan pruning is described in Figure~\ref{fig:trackform:f}, where $N=2$. First, finding the best track hypothesis set is needed before pruning the trees. Computing the best track hypothesis set using the track score is described in Section~\ref{sec:scoring} and ~\ref{sec:besthypo}. After we identify the best hypothesis set, we ascend to the parent node $N$ times from each selected leaf node to find the decision node. At that node, we prune the subtree that diverged from the best track. Consequently, we have a tree of depth $N$ below the decision node, while the tree is degenerated into simple list of assignments above the decision node.

\subsection{Scoring a track}\label{sec:scoring}
The evaluation of a track hypothesis should deal all aspects of data association quality that a multi-camera track possess. According to the original formulation\cite{blackman-book}, we define a likelihood ratio($L$) of a track $T_j=\{o_{I_{j}}\}$ to be 
\begin{equation}\label{eq:score1}
    L(T_j) = \frac{p(\{o_{I_{j}}\}|H_1)}{p(\{o_{I_{j}}\}|H_0)}\frac{P_0(H_1)}{P_0(H_0)}
\end{equation}
where hypotheses $H_1$ and $H_0$ are the true target and false alarm hypotheses of given combination of data, i.e.\ $p(\{o_{I_{j}}\}|H_i)$ is probability density function evaluated with given data $\{o_{I_j}\}$ under the assumption that $H_i$ is correct. The $P_0(H_i)$ is a prior probability of $H_i$. The conditional probabilities in Equation~(\ref{eq:score1}) can be partitioned into a product of two terms, $L_A$ and $L_X$, assuming that the appearance and kinematic information of a target are independent each other. Therefore,
\begin{equation}\label{eq:score2}
\resizebox{0.9\linewidth}{!}{$
    L(T_j) = L_0L_A(T_j)L_X(T_j) = L_0\frac{p(\{A_{I_{j}}\}|H_1)}{p(\{A_{I_{j}}\}|H_0)}\frac{p(\{X_{I_{j}},\pi_{I_j}\}|H_1)}{p(\{X_{I_{j}},\pi_{I_j}\}|H_0)}$}
\end{equation}
where $L_0 = \frac{P_0(H_1)}{P_0(H_0)}$, the second term and third term in rightmost side are $L_A(T_j)$ and $L_X(T_j)$, respectively. The Equation~(\ref{eq:score2}) can be further factorized by chain-rules:
\begin{equation}\label{eq:score3}
    \begin{split}
        L(T_j) &= L_0\prod_{k=1}^{|I_j|}L_A(T_j^k)L_X(T_j^k) \\
        &= \resizebox{0.8\linewidth}{!}{$ L_0\prod_{k=1}^{|I_j|}\dfrac{p(A_{I_{j}^k}|\{A_{I_j^{1:k-1}}\},H_1)}{p(A_{I_{j}^k}|H_0)}\dfrac{p(X_{I_{j}^k},\pi_{I_j^k}|\{X_{I_j^{1:k-1}},\pi_{I_j^{1:k-1}}\},H_1)}{p(X_{I_{j}^k},\pi_{I_j^k}|H_0)}$}
    \end{split}
\end{equation}
where assuming that received observations are conditionally independent under the false alarm hypothesis. $L_A(T_j^k)$ and $L_X(T_j^k)$ are the appearance and kinematic likelihood ratio when the $k$-th observation is associated with the existing track $T_j^{k-1}$.

For likelihood of the kinematic term, $L_X$, we define two different measures in order to differentiate the tracking scenarios. The first one is for the scenario that tracking targets on the image plane of each camera is only available. In this case, we assumed that transition time across cameras is normally distributed, i.e.\
\begin{equation}\label{eq:score4}
\resizebox{0.9\linewidth}{!}{$
    p(X_{I_{j}^k},\pi_{I_j^k}|\{X_{I_j^{1:k-1}},\pi_{I_j^{1:k-1}}\},H_1) = N(t_{I_j^k}^1-t_{I_j^{k-1}}^{|X_{I_j^{k-1}}|};\mu,\sigma^2) $}
\end{equation}
where the mean $\mu$ and variance $\sigma^2$ would be estimated using training samples which moved from $\pi_{I_j^{k-1}}^{ex}$ to $\pi_{I_j^{k}}^{en}$. Note that we dropped sub-script for $\mu$ and $\sigma^2$ for simplicity. They should be learned for all pairs of possible transitions (i.e., for all $\Phi_{e_i,e_j}=1$ where $e_i, e_j \in E$). The $t_{I_j^k}^1$ is the time stamp of the initiation time of $o_{I_j^k}$ whereas $t_{I_j^{k-1}}^{|X_{I_j^{k-1}}|}$ is the time stamp of the last observed time of $o_{I_j^{k-1}}$.

The other kinematic likelihood function is for the scenario where we can locate a target on the ground plane of the camera network; hence, measuring distance between tracks is feasible. Let $\dot{y}_i$ be the moving speed of an observation $o_i$ and it is estimated by averaging:
\begin{equation}\label{eq:score5}
    \dot{y}_i = \frac{1}{|X_i|-1}\sum_{l=2}^{|X_i|}\frac{\|\mathbf{y}_i^l-\mathbf{y}_i^{l-1}\|_2}{t_i^l-t_i^{l-1}}.
\end{equation}
Then the likelihood function is also assumed to be Gaussian:
\begin{equation}\label{eq:score6}
    \begin{split}
        &p(X_{I_{j}^k},\pi_{I_j^k}|\{X_{I_j^{1:k-1}},\pi_{I_j^{1:k-1}}\},H_1) = N(\hat{d}-d;0,\dot{y}_i/\gamma) \\
        &\hat{d}=\dot{y}_{I_j^{k-1}}(t_{I_j^k}^1-t_{I_j^{k-1}}^{|X_{I_j^{k-1}}|}) \\
        &d=\beta\|\mathbf{y}_{I_j^{k}}^1-\mathbf{y}_{I_j^{k-1}}^{|X_{I_j^{k-1}}|}\|_2 + (1-\beta)\|\mathbf{y}_{I_j^{k}}^1-\mathbf{y}_{I_j^{k-1}}^{|X_{I_j^{k-1}}|}\|_1,
    \end{split}
\end{equation}
where $\hat{d}$ is the estimated travel distance of $o_{I_j^{k-1}}$, and $d$ (which comes from the Equation~(\ref{eq:speedgating})) is distance between $o_{I_j^{k-1}}$ and $o_{I_j^{k}}$. Note that although both $\hat{d}$ and $d$ are function of $X_{I_{j}^k}$ and $X_{I_{j}^{k-1}}$ we dropped them for the simplicity of notation. The $\gamma$ is the precision for the Gaussian distribution. For the false alarm hypothesis of kinematic term, $p(X_{I_{j}^k},\pi_{I_j^k}|H_0)$, we set to constant probability, $0< C_1 < 1$. 

To compute the appearance likelihood, we first built an color histogram for the appearance feature $A_i$ of observation $o_i$ while it has tracked. The learned appearance model, $\bar{A}_{I_j^{1:k}}$, is constructed for the track $T_j^{1:k} = \{o_{I_j^{l}}\}_{l=1:k}$, after each association between the existing track $T_j^{1:k-1}$ and a new observation $o_{I_j^k}$ is made, i.e.\:
\begin{equation}\label{eq:score7}
    \bar{A}_{I_j^{1:k}}= \frac{k-1}{k}\bar{A}_{I_j^{1:k-1}}+\frac{1}{k}A_{I_j^k}
\end{equation}
where $\bar{A}_{I_j^{1:k-1}}$ is learned appearance feature for the track $T_j^{1:k-1}$. Thus, $\bar{A}_{I_j^{1:k}}$ is the averaged feature over $k$ associated observations. This averaging model is used because if the track hypothesis consistently associated the observations which had the same identity then the averaged feature would have the ability to classify correctly than that of inconsistently associated track due to its stable distribution of colors. Then the appearance likelihood is computed by comparing two histogram:
\begin{equation}\label{eq:score8}
\resizebox{0.9\linewidth}{!}{$
    p(A_{I_{j}^k}|\{A_{I_j^{1:k-1}}\},H_1) = p(A_{I_{j}^k}|\bar{A}_{I_j^{1:k-1}},H_1) = D_h(A_{I_{j}^k},\bar{A}_{I_j^{1:k-1}}), $}
\end{equation}
where $D_h$ is similarity measure between two histograms and it can be any metric such as, Bhattacharyya, histogram intersection, earth mover distance and so on. However, some metrics should be modified in order to use it as the probability (i.e., $0 \leq D_h \leq 1$). The false alarm hypothesis of appearance term, $p(A_{I_{j}^k}|H_0)$, is the constant probability $0<C_2<1$.

Next, we define the log likelihood ratio, or score, for a multi-camera track, $T_j=\{o_{I_j}\}$ consists of $|I_j|$ observations, which is the sum of $|I_j|$ kinematics and $|I_j|$ appearance related terms plus the initiation score. That is:
\begin{equation}
    \begin{split}
        \log{L(T_j)} &= \log{L_0} + \sum_{k=1}^{|I_j|}[\log{L_A(T_j^k)} + \log{L_X(T_j^k)}]
    \end{split}
\end{equation}
where $\log{L_0}$ is track initiation score and we set it to a constant $C_0$. Then the score of a track can be computed recursively\cite{blackman-book}:
\begin{equation}
    \begin{split}
        \log{L(T_j)} & = \log{L(T_j^{1:|I_j|})} \\
        &= \resizebox{0.65\linewidth}{!}{$ \log{L_0} + \sum_{k=1}^{|I_j|-1}[\log{L_A(T_j^k)} + \log{L_X(T_j^k)}] $} \\
        & \resizebox{0.55\linewidth}{!}{$ \quad+ [\log{L_A(T_j^{|I_j|})} + \log{L_X(T_j^{|I_j|})}] $} \\
        & = \resizebox{0.55\linewidth}{!}{$ \log{L(T_j^{1:|I_j|-1})} + \Delta\log{L(T_j^{|I_j|})}, $}
    \end{split}
\end{equation}

where $\log{L(T_j^{1:|I_j|-1})}$ is the score of track $T_j^{1:|I_j|-1}$ and $\Delta\log{L(T_j^{|I_j|})}$ is the increment that occurs upon update with a new observation, $o_{I_j^{|I_j|}}$. Finally, we introduce the weights, $w_A$ and $w_X$, that control contribution of appearance and kinematics to the score, respectively:
\begin{equation}
    \begin{split}
        \resizebox{0.2\linewidth}{!}{$ \log{L(T_j^{1:|I_j|-1})} $} &=\resizebox{0.7\linewidth}{!}{$ \log{L_0} + \sum_{k=1}^{|I_j|-1}[w_A\log{L_A(T_j^k)} + w_X\log{L_X(T_j^k)}] $} \\
        \resizebox{0.2\linewidth}{!}{$ \Delta\log{L(T_j^{|I_j|})} $} &= \resizebox{0.6\linewidth}{!}{$ w_A\log{L_A(T_j^{|I_j|})} + w_X\log{L_X(T_j^{|I_j|})} $}
    \end{split}
\end{equation}
where $w_A+w_X=1$. The score is continuously updated as long as the track hypothesis is updated with a new observation.

\begin{SCfigure}[1.3][t]
\includegraphics[width=0.38\linewidth]{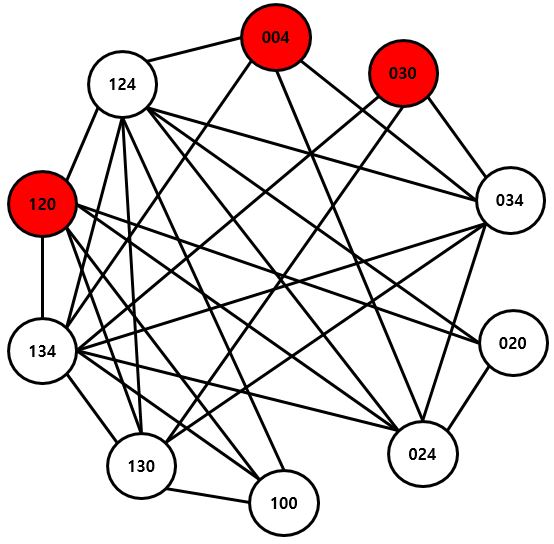}
\caption{An example of MWIS graph that corresponds to the trees in Figure~\protect\ref{fig:trackform:e}. The red vertices have been selected for the best hypothesis set.}
\label{fig:mwis}
\end{SCfigure}

\subsection{Computing the Best Hypothesis Set}\label{sec:besthypo}

In this section, we describe how the best hypothesis set is computed among all track-hypothesis trees that maintain all the possible multi-camera tracks using all observations. The best hypothesis set is computed every scan if the scan received any new observation, and then tree pruning is performed to avoid exponential growth of trees.

To compute the best hypothesis set, we adopt the approach of~\cite{mht-rv}. They change the task to the Maximum Weighted Independent Set (MWIS) problem. MWIS\cite{papageorgiou2009maximum} is equivalent to the Multi-Dimensional Assignment (MDA) problem in the context of MHT\cite{mht-rv}. MDA is used to compute the most probable set of tracks\cite{blackman-book,poore1993data} and its MDA formulation for MHT was introduced in~\cite{blackman-book,poore1993data}.

Let $\mathcal{G}=(\mathcal{Z},\mathcal{E})$ be the undirected graph for MWIS, which corresponds to a set of track-hypothesis trees generated by MHT. Then the solution of MWIS can be determined by solving the discrete optimization problem:
\begin{equation}\label{eq:best1}
    \begin{split}
        & \max_{\mathbf{z}}{\sum_{i}{c_iz_i}} \\
        \text{s.t.}\quad z_i+z_j & \leq 1, \quad \forall (i,j)\in \mathcal{E},\quad z_i\in \{0,1\}.
    \end{split}
\end{equation}
Each vertex $z_i\in \mathcal{Z}$ is assigned to a track hypothesis $T_i$ as well as a vertex $z_i$ has weight $c_i$ that corresponds to its track score, $\log{L(T_i)}$. There is an undirected edge, $(i,j)\in \mathcal{E}$, linking two vertices $z_i$ and $z_j$ if the two tracks are incompatible due to shared observations, i.e.\ if they are violating constraint~(\ref{eq:set-constraint}). Therefore, the constraint in Equation~(\ref{eq:best1}) is a discretized form of constraint~(\ref{eq:set-constraint}). In the graph $\mathcal{G}$, an independence set is a set of vertices no two of which are adjacent, i.e.\ all tracks in the independence set are compatible. The maximum weighted independence set in $\mathcal{G}$ is an independence set with maximum total weight, i.e.\ a set of compatible tracks of which the total track score is maxima. An example of the MWIS graph is shown in Figure~\ref{fig:mwis}. It corresponds to the set of track-hypothesis trees in Figure~\ref{fig:trackform:e}. Each vertex represents a track hypothesis(each branch of a tree) and the observations used for that track are shown at the vertex with three-digits, where zero denotes a dummy observation. Note that since three scans(first, third and fifth scans) have received observations (among the six scans in Figure~\ref{fig:trackform:a}) a track can contain at most three observations(Refer to Section~\ref{sec:mht-tree} for more details). The red vertices are selected for the best hypothesis set in the example(Figure~\ref{fig:mwis}). We used the Gurobi optimizer to solve the above MWIS problem.

\section{Experiments}\label{sec:exp}
We evaluated our method using two datasets: DukeMTMC \cite{ristani2016MTMC} and NLPR\_MCT \cite{mct2014}. These datasets were designed for a multi-camera tracking system. The DukeMTMC dataset consists of eight synchronized cameras, which was recorded at 1080p resolution and 60 fps. The dataset contains more than 7,000 single camera trajectories and over 2,000 unique identities over $85$ minutes for each camera, a total of more than 10 h. We used \textit{ID-Measure} \cite{ristani2016MTMC} with the DukeMTMC dataset to evaluate our multi-camera tracking performance. The NLPR\_MCT dataset provides four different videos that are, at most, $25$ minutes long, and they all have a resolution of $320\times 240$. To measure the performance for this dataset, we used the \textit{MCTA} metric\cite{mct2014}. Table \ref{exp:params} shows the parameter setting for each dataset in this section. Note that $G_{speed}$ is in meters per second and $G_{time}$ is in seconds.
\begin{table*}[ht]
    \centering
    \small
    \resizebox{0.75\textwidth}{!}{\begin{tabular}{c|c|c|c|c|c|c|c|c}
        \hline
        dataset & $N$-pruning & $w_A$ & $C_0$ & $C_1$ & $C_2$ & scan time & $\beta, G_{speed}^{min}, G_{speed}^{max}$ & $G_{time}^{min}, G_{time}^{max}$ \\
        \hline
        \hline
        DukeMTMC  & 10 & 0.8  & 0.001 & 0.3 & 0.75 & 1 sec & 0.7, 0.5, 2.0 & N/A \\
        NLPR\_MCT & 10 & 0.815 & 0.005 & 0.1 & 0.75 & 1 sec & N/A & $\mu-2.5\sigma$, $\mu+2.5\sigma$ \\
        \hline
    \end{tabular}}
    \caption{Parameter settings}
    \label{exp:params}
\end{table*}

To solve the SCT problem as well as to generate observations that is inputted to the proposed MHT, we used online and real-time MOT\cite{ristani2014tracking} for each camera. Utilizing the online and real-time single camera tracker would not affect the online and real-time capability of unified framework. In Section \ref{sec:realtime}, we show that our unified framework works in real-time provided that the SCT problem is solved in real-time.

\subsection{Appearance Modeling}

\begin{figure}[t]
    \centering
    \subfloat[]{\label{fig:CPM_Model:a}\includegraphics[height=2.5cm]{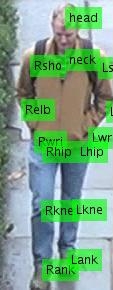}  \includegraphics[height=2.5cm]{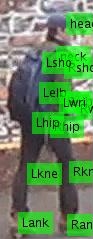}}
    \qquad
    \subfloat[]{\label{fig:CPM_Model:b}\includegraphics[height=2.5cm]{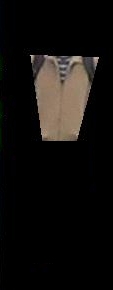}  \includegraphics[height=2.5cm]{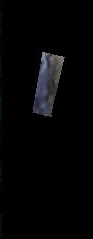}}
    \qquad
    \subfloat[]{\label{fig:CPM_Model:c}\includegraphics[height=2.5cm]{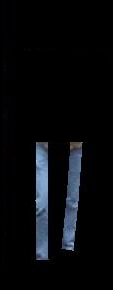}  \includegraphics[height=2.5cm]{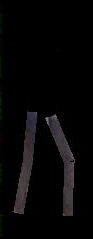}}
    
    \caption{\protect\subref{fig:CPM_Model:a} shows the image patches of a person and results of pose estimation. \protect\subref{fig:CPM_Model:b}: corresponding upper body parts. \protect\subref{fig:CPM_Model:c}: corresponding bottom body parts.}
    \label{fig:CPM_Model}
\end{figure}

In this subsection, the appearance model for an observation is described. The appearance feature $A_i$ is the averaged histogram of observation $o_i$ that is learned while under tracking status($w_1$) of $o_i$. To be more specific, let $\mathbf{a}_i^l$ be the extracted feature from corresponding image location $\mathbf{x}_i^l$, which is the $l$-th track histroy of observation $i$. Then the appearance feature $A_i^l$ that is computed from start to the $l$-th track history is:

\begin{equation}
    A_i^l = \frac{l-1}{l}A_i^{l-1} + \frac{1}{l}\mathbf{a}_i^l
\end{equation}
where $l \leq |X_i|$. Note that to keep the online nature of the proposed method, $A_i^l$ is used to compute Equation (\ref{eq:score8}) instead of $A_i$. 

For appearance feature, we use an HSV (hue, saturation, value) color histogram for upper and lower body parts where the bin size is 16, 4 and 4 for Hue, Saturation and Value channels respectively. Furthermore, to capture the pose variations of a person, we use the Convolutional Pose Machine\cite{wei2016cpm} to estimate the pose of a given image patch of a person. The estimated pose of a person is depicted in Figure \ref{fig:CPM_Model:a}. To extract the upper body part, four joints(right shoulder, right hip, left shoulder, left hip) are used(Figure \ref{fig:CPM_Model:b}). Six joints(right hip, right knee, right ankle, left hip, left knee, left ankle) are used for the bottom part of the body (Figure \ref{fig:CPM_Model:c}). Once those body parts are extracted from an image patch, an HSV histogram, $\mathbf{a}_i^l$, is computed.

Even though we have used color histogram as the appearance feature and simple averaging model in this work, other online appearnce model can be applied to proposed MHT method since it is known that MHT can be extended by including online learned discriminative models without difficulty \cite{mht-rv}.

\subsection{DukeMTMC dataset}

\begin{table*}[t]
\centering
\resizebox{0.8\textwidth}{5cm}{\begin{tabular}{c|c|ccc|cccccccc}
\hline
\multirow{2}{*}{Cam} & \multirow{2}{*}{method} & \multicolumn{3}{c|}{ID-Measure} & \multicolumn{8}{c}{CLEAR MOT}\\
 & & IDF1$\uparrow$ & IDP$\uparrow$ & IDR$\uparrow$ & MOTA$\uparrow$ & MOTP$\uparrow$ & FAF$\downarrow$ & MT$\uparrow$ & ML$\downarrow$ & FP$\downarrow$ & FN$\downarrow$ & IDs$\downarrow$ \\
\hline
\hline
\multirow{3}{*}{1}
 & \cite{ristani2016MTMC} & 57.3 & 91.2 & 41.8 & 43.0 & 79.0 & 0.03 & 24 & 46 & 2,713 & 107,178 & 39\\
 & \cite{tesfaye2017multi} & 76.9 & 89.1 & 67.7 & 69.9 & 76.3 & 0.06 & 137 & 22 & 5,809 & 52,152 & 156 \\
 & Ours & 84.3 & 89.7 & 79.6 & 84.9 & 79.5 & 0.04 & 191 & 12 & 3,679 & 25,318 &55 \\
\hline
\multirow{3}{*}{2}
 & \cite{ristani2016MTMC} & 68.0 & 69.3 & 67.1 & 44.8 & 78.2 & 0.51 & 133 & 8 & 47,919 & 5,374 & 60\\
 & \cite{tesfaye2017multi} & 81.2 & 90.9 & 73.4 & 71.5 & 74.6 & 0.09 & 134 & 21 & 8,487 & 43,912 & 75\\
 & Ours & 81.9 & 88.9 & 75.9 & 78.4 & 77.1 & 0.07 & 151 & 8 & 6,390 & 33,377 & 81\\
\hline
\multirow{3}{*}{3}
 & \cite{ristani2016MTMC} & 60.3 & 78.9 & 48.8 & 57.8 & 77.5 & 0.02 & 52 & 22 & 1,438 & 28,692 & 16\\
 & \cite{tesfaye2017multi} & 64.6 & 76.3 & 56.0 & 67.4 & 75.6 & 0.02 & 44 & 9 & 2,148 & 21,125 & 38 \\
 & Ours & 69.3 & 76.2 & 63.5 & 65.7 & 77.0 & 0.06 & 58 & 7 & 5,908 & 18,589 & 22\\
\hline
\multirow{3}{*}{4}
 & \cite{ristani2016MTMC} & 73.5 & 88.7 & 62.8 & 63.2 & 80.2 & 0.02 & 36 & 18 & 2,209 & 19,323 & 7\\
 & \cite{tesfaye2017multi} & 84.7 & 91.2 & 79.0 & 76.8 & 76.6 & 0.03 & 45 & 4 & 2,860 & 10,686 & 18\\
 & Ours & 80.7 & 84.1 & 77.6 & 79.8 & 80.1 & 0.04 & 47 & 3 & 3,633 & 8,173 & 17\\
\hline
\multirow{3}{*}{5}
 & \cite{ristani2016MTMC} & 73.2 & 83.0 & 65.4 & 72.8 & 80.4 & 0.05 & 107 & 17 & 4,464 & 35,861 & 54\\
 & \cite{tesfaye2017multi} & 68.3 & 76.1 & 61.9 & 68.9 & 77.4 & 0.10 & 88 & 11 & 9,117 & 36,933 & 139\\
 & Ours & 73.7 & 81.4 & 67.3 & 76.6 & 80.0 & 0.05 & 110 & 8 & 4,410 & 30,195 & 83\\
\hline
\multirow{3}{*}{6}
 & \cite{ristani2016MTMC} & 77.2 & 84.5 & 69.1 & 73.4 & 80.2 & 0.06 & 142 & 27 & 5,279 & 45,170 & 55\\
 & \cite{tesfaye2017multi} & 82.7 & 91.6 & 75.3 & 77.0 & 77.2 & 0.05 & 136 & 11 & 4,868 & 38,611 & 142\\
 & Ours & 83.5 & 88.9 & 78.8 & 82.8 & 80.2 & 0.06 & 163 & 6 & 5,478 & 27,194 & 69\\
\hline
\multirow{3}{*}{7}
 & \cite{ristani2016MTMC} & 80.5 & 93.6 & 70.6 & 71.4 & 74.7 & 0.02 & 69 & 13 & 1,395 & 18,904 & 23\\
 & \cite{tesfaye2017multi} & 81.6 & 94.0 & 72.5 & 73.8 & 74.0 & 0.01 & 64 & 4 & 1,182 & 17,411 & 36\\
 & Ours & 81.5 & 91.4 & 73.5 & 77.0 & 75.5 & 0.01 & 69 & 7 & 1,232 & 15,119 & 33\\
\hline
\multirow{3}{*}{8}
 & \cite{ristani2016MTMC} & 72.4 & 92.2 & 59.6 & 60.7 & 76.7 & 0.03 & 102 & 53 & 2,730 & 52,806 & 46\\
 & \cite{tesfaye2017multi} & 73.0 & 89.1 & 61.0 & 63.4 & 73.6 & 0.04 & 92 & 28 & 4,184 & 47,565 & 91 \\
 & Ours & 79.9 & 90.8 & 71.3 & 71.6 & 75.3 & 0.05 & 125 & 21 & 4,850 & 35,288 & 46\\
\hline
\multirow{3}{*}{ \makecell{Single Cam\\ Average} }
 & \cite{ristani2016MTMC} & 70.1 & 83.6 & 60.4 & 59.4 & 78.7 & 0.09 & 665 & 234 & 68,147 & 361,672 & 91\\
 & \cite{tesfaye2017multi} & 77.0 & \textbf{87.6} & 68.6 & 70.9 & 75.8 & 0.05 & 740 & 110 & 38,655 & 268,398 & 693\\
 & Ours & \textbf{80.3} & 87.3 & \textbf{74.4} & 78.3 & 78.4 & 0.05 & 914 & 72 & 35,580 & 193,253 & 406\\
\hline
\hline
\multirow{3}{*}{Multi-Cam}
 & \cite{ristani2016MTMC} & 56.2 & 67.0 & 48.4\\
 & \cite{tesfaye2017multi} & 60.0 & 68.3 & 53.5\\
 & Ours & \textbf{65.4} & \textbf{71.1} & \textbf{60.6}\\
\hline
\end{tabular}}
\caption{This table compares the results for each camera and average performance over all single cameras as well as the performance of multi-camera tracking on the Test\_easy sequence of DukeMTMC dataset.}\label{exp:DukeResult1}
\end{table*}

\begin{figure}[t]
    \centering
    \subfloat[]{\label{fig:DukeMap:a}\includegraphics[width=0.45\linewidth]{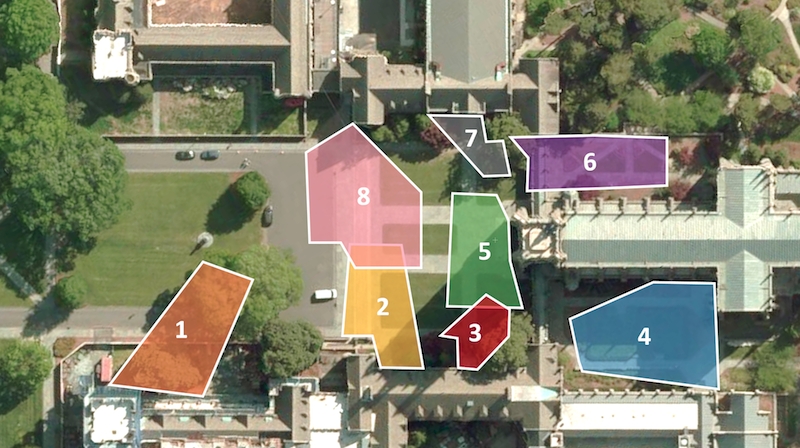}}
    \quad
    \subfloat[]{\label{fig:DukeMap:b}\includegraphics[width=0.42\linewidth]{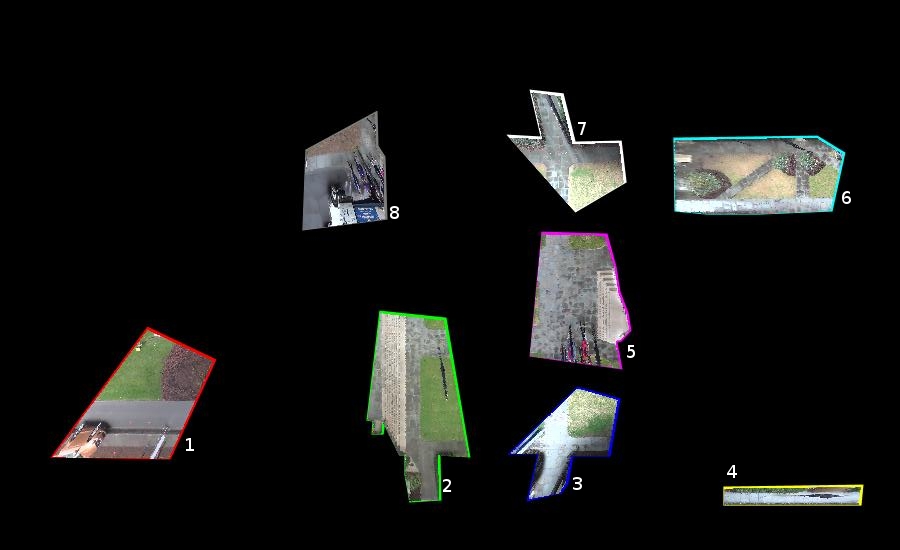}}
    \caption{Camera toplogy of DukeMTMC camera. \protect\subref{fig:DukeMap:a}: Each polygon represents the FoV of corresponding camera. \protect\subref{fig:DukeMap:b}: Each polygon represents the RoI of corresponding camera.}
    \label{fig:DukeMap}
\end{figure}

The DukeMTMC is a large, fully-annotated, calibrated dataset that captures the campus of Duke University, and was recorded using eight fixed cameras. The dataset has an RoI (region of interest) area for each camera where an evaluation is made. The topology of the fixed cameras is shown in Figure \ref{fig:DukeMap} where there is no field overlap between any pair of cameras(Figure \ref{fig:DukeMap:b}). The cameras used to acquire the dataset were synchronized and recorded at 1080p resolution and 60fps. The dataset contains more than 7,000 single camera trajectories and over 2,000 unique identities captures during $85$ minutes of recording for each camera, thus, a total of more than 10 h. The video was split into one training/validation set and two test sets, test-easy and test-hard set. The difficulty of the test-easy set is similar to the training/validation set, and it is 25 minutes long. The test-hard set is  10 minute-long videos and contains a group of dozens of people traversing multiple cameras. 

The evaluation criteria for the DukeMTMC dataset is ID-Measure \cite{ristani2016MTMC}, which measures how well a tracker determine who is where at all times. This criteria has three measures: \textit{IDP (Identification precision)}, \textit{IDR (Identification Recall)}, and \textit{IDF1 (Identification F-Score)}. The IDP (IDR) is the fraction of computed (ground truth) detections that are correctly identified. IDF1 is the ratio of correctly identified detections over the average number of ground-truth and computed detections. This process is different from CLEAR MOT\cite{Bernardin2008}, which reports the amount of incorrect decisions made by a tracker. Moreover unlike CLEAR MOT, it can measure not only the single camera tracking results but also the multi-camera tracking results. We reported the single camera tracking performance with both ID-Measure and CLEAR MOT in order to make it clear how much proposed method improved the final tracking performance given these single camera tracks.

We compared the quantitative performance of our method with other multi-target multi-camera tracking methods\cite{ristani2016MTMC, tesfaye2017multi} using the DukeMTMC dataset. The results are shown in Table \ref{exp:DukeResult1} and Table \ref{exp:DukeResult2}. The evaluation on the Test-easy is shown in Table \ref{exp:DukeResult1}, while the performance on the Test-hard is shown in Table \ref{exp:DukeResult2}. In both tables, the last row is for comparison of multi-camera tracking performance and the rest are for comparison of single camera tracking performance. We used public detection responses as the input to our method. The results of single camera tracking between ours and \cite{ristani2016MTMC} were different, even though we used the public single camera tracker that is published by E. Ristani \etal\cite{ristani2014tracking}, because we modified the original one to fit our multi-camera tracking method.\footnote{The modified version of \cite{ristani2014tracking} is currently on the web.\\
https://github.com/yoon28/SCT4DukeMTMC/} 
The last row of Table \ref{exp:DukeResult1} shows that the performance of our multi-camera tracking method outperformed the state-of-the-art method\cite{tesfaye2017multi} on the Test-easy sequence by $5.4\%$ in IDF1, $2.8\%$ in IDP and $7.1\%$ in IDR metrics. In the Test-hard sequence, the proposed method ranked second with difference of $0.8\%$ in IDF1 and $4.9\%$ in IDP metrics, while it was first in IDR metrics (improvement of $1.3\%$, Table \ref{exp:DukeResult2}). Finally, the proposed MHT algorithm for the multi-camera tracking method outperformed the method of \cite{ristani2016MTMC} even in the complicated video sequence(Test-hard). Even if the \cite{ristani2016MTMC}'s average IDF1 over single cameras was higher than ours by $1\%$, the IDF1 of their multi-camera tracking performance was even lower than ours by $2.8\%$(Table \ref{exp:DukeResult2}).

\begin{table*}[t]
\centering
\resizebox{0.8\textwidth}{5cm}{\begin{tabular}{c|c|ccc|cccccccc}
\hline
\multirow{2}{*}{Cam} & \multirow{2}{*}{method} & \multicolumn{3}{c|}{ID-Measure} & \multicolumn{8}{c}{CLEAR MOT}\\
 & & IDF1$\uparrow$ & IDP$\uparrow$ & IDR$\uparrow$ & MOTA$\uparrow$ & MOTP$\uparrow$ & FAF$\downarrow$ & MT$\uparrow$ & ML$\downarrow$ & FP$\downarrow$ & FN$\downarrow$ & IDs$\downarrow$ \\
\hline
\hline
\multirow{3}{*}{1}
 & \cite{ristani2016MTMC} & 52.7 & 92.5 & 36.8 & 37.8 & 78.1 & 0.03 & 6 & 34 & 1,257 & 78,977 & 55\\
 & \cite{tesfaye2017multi} & 67.1 & 83.0 & 56.4 & 63.2 & 75.7 & 0.08 & 65 & 17 & 2,886 & 44,253 & 408\\
 & Ours & 64.6 & 72.2 & 58.4 & 61.1 & 76.7 & 0.35 & 78 & 11 & 12,570 & 37,287 & 394\\
\hline
\multirow{3}{*}{2}
 & \cite{ristani2016MTMC} & 60.6 & 65.7 & 56.1 & 47.3 & 76.5 & 0.74 & 68 & 12 & 26,526 & 46,898 & 194\\
 & \cite{tesfaye2017multi} & 63.4 &78.8 & 53.1 & 54.8 & 73.9 & 0.24 & 62 & 16 & 8,653 & 54,252 & 323\\
 & Ours & 56.6 & 61.2 & 52.6 & 50.4 & 74.4 & 0.68 & 66 & 10 & 24,591 & 44,401 & 392\\
\hline
\multirow{3}{*}{3}
 & \cite{ristani2016MTMC} & 62.7 & 96.1 & 46.5 & 46.7 & 77.9 & 0.01 & 24 & 4 & 288 & 18,182 & 6\\
 & \cite{tesfaye2017multi} & 81.5 & 91.1 & 73.7 & 68.8 & 75.1 & 0.06 & 18 & 2 & 2,093 & 8,701 & 11\\
 & Ours & 80.0 & 86.9 & 74.1 & 70.3 & 76.8 & 0.07 & 22 & 2 & 2,543 & 7,737 & 10\\
\hline
\multirow{3}{*}{4}
 & \cite{ristani2016MTMC} & 84.3 & 86.0 & 82.7 & 85.3 & 81.5 & 0.04 & 21 & 0 & 1,215 & 2,073 & 1\\
 & \cite{tesfaye2017multi} & 82.3 & 87.1 & 78.1 & 75.6 & 77.7 & 0.05 & 17 & 0 & 1,571 & 3,888 & 61\\
 & Ours & 83.3 & 84.4 & 82.2 & 81.2 & 81.6 & 0.05 & 20 & 1 & 1,821 & 2,404 & 1\\
\hline
\multirow{3}{*}{5}
 & \cite{ristani2016MTMC} & 81.9 & 90.1 & 75.1 & 78.3 & 80.7 & 0.04 & 57 & 2 & 1,480 & 11,568 & 13\\
 & \cite{tesfaye2017multi} & 82.8 & 91.5 & 75.7 & 78.6 & 76.7 & 0.03 & 47 & 2 & 1,219 & 11,644 & 50\\
 & Ours & 85.7 & 93.3 & 79.2 & 81.9 & 80.1 & 0.02 & 52 & 2 & 875 & 10,017 & 24\\
\hline
\multirow{3}{*}{6}
 & \cite{ristani2016MTMC} & 64.1 & 81.7 & 52.7 & 59.4 & 76.7 & 0.14 & 85 & 23 & 5,156 & 77,031 & 225\\
 & \cite{tesfaye2017multi} & 53.1 & 71.2 & 42.3 & 53.3 & 76.5 & 0.17 & 68 & 36 & 5,989 & 88,164 & 547\\
 & Ours & 54.7 & 70.0 & 44.9 & 56.1 & 77.8 & 0.22 & 82 & 24 & 7,902 & 80,716 & 423\\
\hline
\multirow{3}{*}{7}
 & \cite{ristani2016MTMC} & 59.6 & 81.2 & 47.1 & 50.8 & 73.3 & 0.08 & 43 & 23 & 2,971 & 38,912 & 148\\
 & \cite{tesfaye2017multi} & 60.6 & 84.7 & 47.1 & 50.8 & 74.0 & 0.05 & 34 & 20 & 1,935 & 39,865 & 266\\
 & Ours & 55.7 & 74.7 & 44.4 & 49.8 & 73.8 & 0.11 & 42 & 25 & 4,405 & 38,687 & 214\\
\hline
\multirow{3}{*}{8}
 & \cite{ristani2016MTMC} & 82.4 & 94.9 & 72.8 & 73.0 & 75.9 & 0.02 & 34 & 5 & 706 & 9,735 & 10\\
 & \cite{tesfaye2017multi} & 81.3 & 90.3 & 73.9 & 70.0 & 72.6 & 0.06 & 37 & 6 & 2,297 & 9,306 & 26\\
 & Ours & 80.4 & 93.5 & 70.5 & 71.5 & 74.0 & 0.02 & 38 & 5 & 731 & 10,278 & 10\\
\hline
\multirow{3}{*}{ \makecell{Single Cam\\ Average} }
 & \cite{ristani2016MTMC} & 64.5 & 81.2 & 53.5 & 54.6 & 77.1 & 0.14 & 338 & 103 & 39,599 & 283,376 & 652\\
 & \cite{tesfaye2017multi} & \textbf{65.4} & \textbf{81.4} & 54.7 & 59.6 & 75.4 & 0.09 & 348 & 99 & 26,643 & 260,073 & 1637\\
 & Ours & 63.5 & 73.9 & \textbf{55.6} & 59.6 & 76.7 & 0.19 & 400 & 80 & 55,038 & 231,527 & 1468\\
\hline
\hline
\multirow{3}{*}{Multi-Cam}
 & \cite{ristani2016MTMC} & 47.3 & 59.6 & 39.2\\
 & \cite{tesfaye2017multi} & \textbf{50.9} & \textbf{63.2} & 42.6\\
 & Ours & 50.1 & 58.3 & \textbf{43.9}\\
\hline
\end{tabular}}
\caption{This table compares the results for each camera and average performance over all single cameras as well as the performance of multi-camera tracking on the Test\_hard sequence of DukeMTMC dataset.}\label{exp:DukeResult2}
\end{table*}

\subsection{NLPR\_MCT dataset}

\begin{figure}[t]
    \centering
    \includegraphics[width=0.9\linewidth]{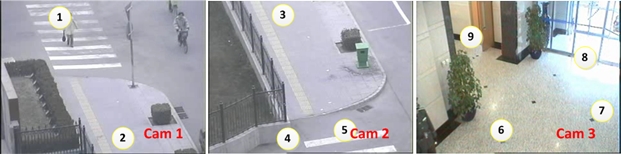}
    \caption{A sub-dataset of NLPR\_MCT datast. The numbers in each scene represent the entry/exit points.}
    \label{fig:NLPR_example}
\end{figure}

The NLPR\_MCT dataset consists of four sub-datasets. A sub-dataset is depicted in Figure \ref{fig:NLPR_example}\footnote{This figure is captured from http://mct.idealtest.org/}. Each sub-dataset includes 3-5 cameras with non-overlapping scenes and recordes different situations according to the number of people (ranging from 14 to 255) and the level of illumination changes and occlusions\cite{mct2014}. The videos contain both real scenes and simulated environments. Each video was nearly 20 minutes long (except Dataset 3), with a rate of 25 fps. 

In this dataset, the topological connection information for every pair of entry/exit points for each sub-dataset is provided. We split the $\pi_i$ of an observation $o_i$ into $\pi_i^{en}$ and $\pi_i^{ex}$, that represent the entry point and exit point of observation $o_i$, respectively. Because the dataset did not provide separate training and test datasets, we learned the parameters for our method as well as the transition matrix, the mean and standard deviation of transition time for each possible transition pair of entry/exit points using first 70 percent of each dataset.

The evaluation criteria used for the NLPR\_MCT dataset was MCTA \cite{mct2014}, multi-camera object tracking accuracy. It was modified based on CLEAR MOT \cite{Bernardin2008} and can be applied to MCT. The metric contains three terms (detection ability, single camera tracking ability and MCT ability), which are multiplied to produce one measure. In this experiment, we used annotated single camera trajectories by assuming that SCT problem is solved in advance\footnote{ This setting is identical to Experiment $1$ of MCT challenge. }. We compared the performance of our method with the state-of-the-art methods in Table \ref{exp:NLPR}. The last column, Avg. Rank, is the averaged ranking over four sub-datasets, where the rank was decided by the MCTA score. This criterion is also used in MCT challenge to compare the results with others. The first place for each sub-dataset was shown in boldface. As a result, both \cite{cai2014exploring} and our method tied for second place by the Avg. Rank of $3.5$. However, it is worth noticing that our method has more stable performance than that of \cite{cai2014exploring} because the rank standard deviation of our method over all sub-datasets is $1$ while the standard deviation of \cite{cai2014exploring} is $1.9149$. Again, the MCTA standard deviation of our method over all sub-datasets is $0.1320$ while that of \cite{cai2014exploring} is $0.1915$.

\begin{table}[t]
\centering
\normalsize
\resizebox{\linewidth}{!}{\begin{tabular}{c|cccc|c}
\hline
Method & NLPR 1 & NLPR 2 & NLPR 3 & NLPR 4 & Avg. Rank \\
\hline \hline
USC\_Vision\cite{cai2014exploring} & 0.9152 & 0.9132 & 0.5162 & 0.7051 & 3.5  \\
UW\_IPL\cite{lee2017online}     & \textbf{0.9610} & \textbf{0.9265} & \textbf{0.7889} & 0.7578 & \textbf{1.25} \\
CRF\_UCR\cite{chen2016integrating}    & 0.8383 & 0.8015 & 0.6645 & 0.7266 & 3.75 \\
EGTracker\cite{mct2014}   & 0.8353 & 0.7034 & 0.7417 & 0.3845 &  4.75 \\
DukeMTMC\cite{ristani2016MTMC}    & 0.7967 & 0.7336 & 0.6543 & \textbf{0.7616} &  4.25 \\ 
Ours        & 0.9129 & 0.8944 & 0.6699 & 0.6812 &  3.5 \\
\hline
\end{tabular}}
\caption{This table compares the results of each sub-dataset. MCTA metric is \\ used for evaluation.}\label{exp:NLPR}
\end{table}

\subsection{N-scan Pruning}

\begin{figure}[t]
    \centering
    \includegraphics[width=0.8\linewidth]{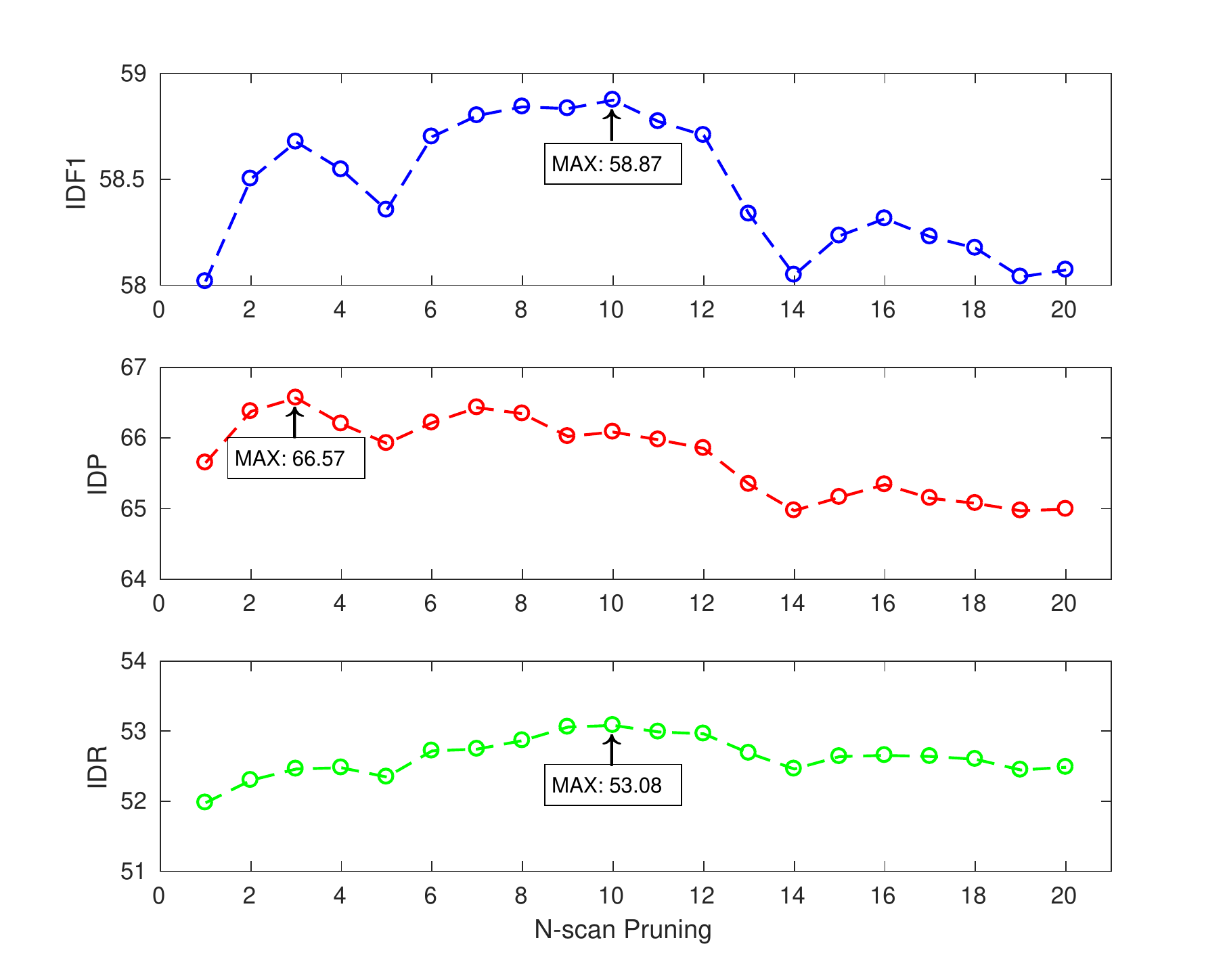}
    \caption{The paramemter $N$ of $N$-scan pruning versus performance of ID-Measure}
    \label{exp:N-pruning}
\end{figure}

In this section, we report how our MHT algorithm is sensitive to the parameter $N$ of $N$-scan pruning. The $N$-scan pruning algorithm assumes that any ambiguity at $K$ is resolved by time $K+N$. The $N$-scan pruning was utilized in the multi-scan assignment approach to MHT because it solves the data association problems with recent $N$ scans of the data thanks to $N$-scan pruning\cite{blackman-book, poore1993data}. We evaluated the ID-Measure for various $N$, $1 \leq N \leq 20$, on the Trainval-Mini of the DukeMTMC dataset. The Trainval-Mini is a small part of the training/validation set of the DukeMTMC dataset and is about 18 minutes long sequence. The parameter settings were the same as before except the $N$. The intersection-over-union is fixed to $0.5$ for this experiment. The experimental result is shown in Figure \ref{exp:N-pruning}. The result demonstrates that our MHT algorithm is negatively affected by $N$ in IDP while it is slightly positively affected in IDR when $N$ is increasing. Therefore, the proposed method has a small sensitivity to $N$ in terms of IDF1 because the IDP and the IDR are negatively correlated with respect to the $N$. The difference between the minimum(58.02\%) and maximum(58.87\%) values in IDF1 was 0.85\%. The minimum and maximum value of IDP was 64.97\% and 66.57\%, respectively, while that of IDR was 51.98\% and 53.08\%, respectively.

\subsection{Real-time implementation}\label{sec:realtime}

\begin{figure}[t]
\centering
\subfloat[camera 1]{\label{fig:gistdata:cam1}\includegraphics[height=0.23\linewidth]{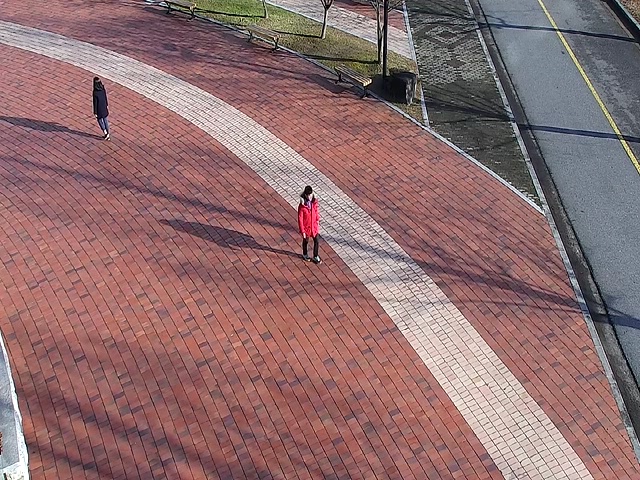}} \quad
\subfloat[camera 2]{\label{fig:gistdata:cam2}\includegraphics[height=0.23\linewidth]{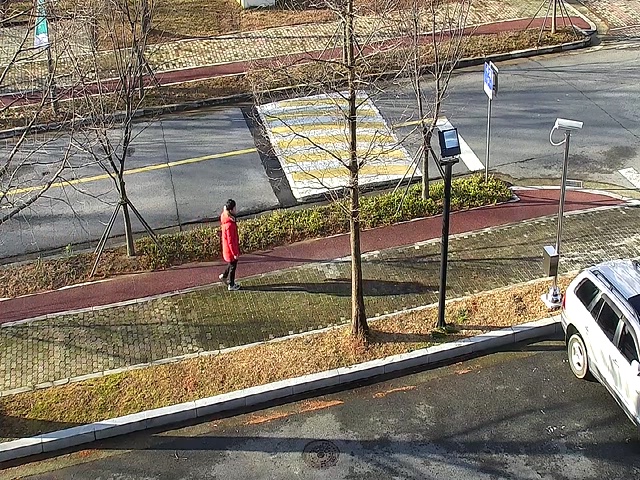}} \quad
\subfloat[camera 3]{\label{fig:gistdata:cam3}\includegraphics[height=0.23\linewidth]{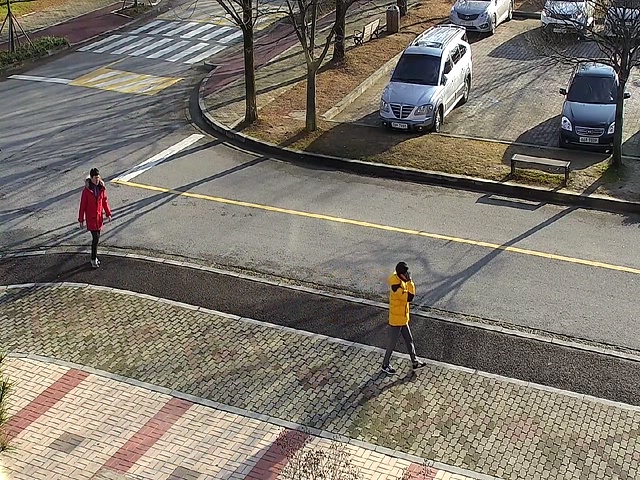}}

\subfloat[camera 4]{\label{fig:gistdata:cam4}\includegraphics[height=0.23\linewidth]{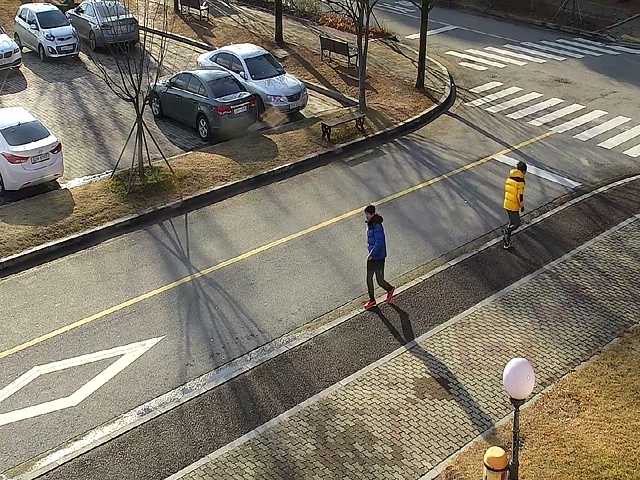}} \quad
\subfloat[camera 5]{\label{fig:gistdata:cam5}\includegraphics[height=0.23\linewidth]{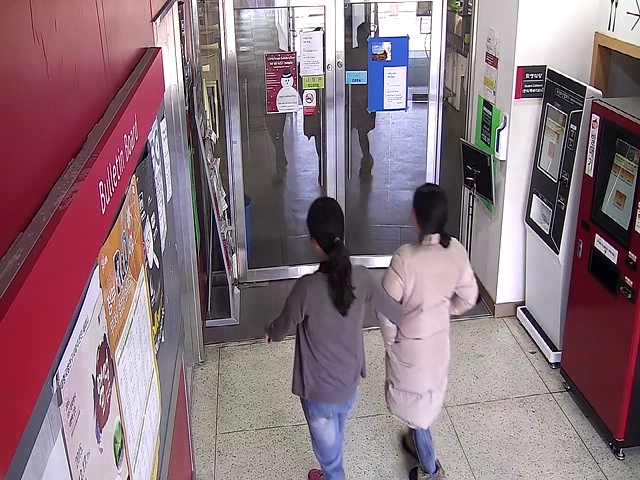}} \quad
\subfloat[camera 6]{\label{fig:gistdata:cam6}\includegraphics[height=0.23\linewidth]{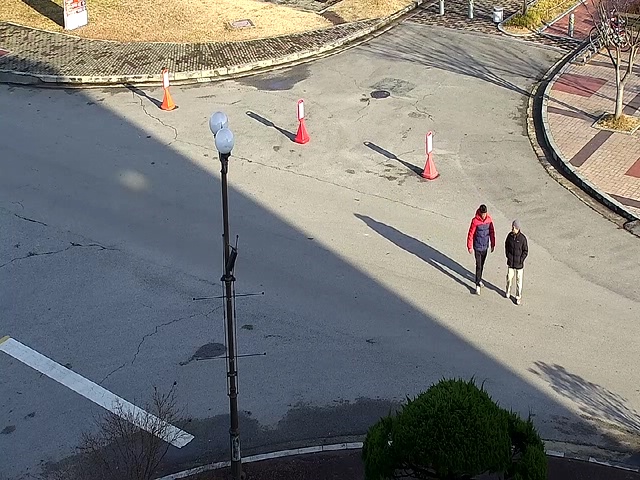}}
\caption{The dataset used in Section \ref{sec:realtime}.}
\label{fig:gistdata}
\end{figure}

In this section, a real-time implementation of the proposed method is described. Full implementation of the proposed method was programmed by Matlab with a desktop PC(Intel i7-4790K 4.0 Ghz 4-core CPU, 16GB RAM, Nvidia GTX 770 GPU and Ubuntu 16.04 OS). For the computation efficiency, we have discarded the Convolutional Pose Machine in the appearance modeling and switched the SCT algorithm from \cite{ristani2014tracking} to GM-PHD(Gaussian Mixture Probability Hypothesis Density) filter\cite{song2016online} for real-time implementation. There was no other reasons for switching the SCT algorithm except that we had already implemented the GMPHD in C++. Real-time implementation was developed using Visual C++ with the multi-thread programming and OpenCV in Windows 10 OS. The test hardware environment included a PC with Intel i7-7700K 4.5 Ghz 4-core CPU, 32GB RAM. To test the processing speed, we have generated a new dataset consisting of six videos with $640\times480$ resolution and about 7 minutes long(Figure \ref{fig:gistdata}). This dataset, which includes appearance of up to 25 targets, was recorded in the campus of Gwangju Institute of Science and Technology. To detect a person in the dataset, we applied the pedestrian detector\cite{kudet}, which processes every frame of each camera to detect pedestrians. The average processing time (includes the processing time of detection, SCT and MCT) was about 15 frames per second for all videos. Note that each video(camera) was processed in parallel by multi-thread programming. Therefore, this result demonstrates the real-time performance of our method. We used the Gurobi optimizer to solve MWIS problems in this implementations.

\section{Conclusion}\label{sec:conclude}

In this paper, we applied a multiple hypothesis tracking algorithm to handle the multi-target multi-camera tracking problem with disjoint views. Our method forms track-hypothesis trees whose branch represents a multi-camera track which describes the trajectory of a target that may move within a camera as well as move across cameras. Furthermore, tracking targets within a camera is performed simultaneously with the tree formation by manipulating a status of each track hypothesis. Besides, two gating schemes have been proposed to differentiate the tracking scenarios. The experimental results shows that our method achieves state-of-the-art performance on DukeMTMC dataset and performs comparable to the state-of-the-art method on NLPR\_MCT dataset. We also show that the proposed method can solve the problem under online and real-time conditions, provided that the single camera tracker solves in such conditions as well.

MHT can be extended by including online learned discriminative appearance models for each track hypothesis\cite{mht-rv}. Therefore, as for the future work, we will investigate online learning techniques that could learn a model for each hypothesis since we used a simple averaging model for appearance modeling in this work. 

\section{Acknowledgement}
This work was supported by Institute for Information and communications Technology Promotion(IITP) grant funded by the Korea government(MSIP) (No. B0101-15-0525, Development of global multi-target tracking and event prediction techniques based on real-time large-scale video analysis). In addition, this work also supported by Korea Creative Content Agency (KOCCA) and Ministry of Culture, Sports and Tourism (MCST) (No. R2017050052, Developed intelligent UI/UX technology for AR glasses-based docent operation).

\bibliographystyle{ieeetr}

\end{document}